\documentclass[10pt,twocolumn,letterpaper]{article}

\usepackage{cvpr}
\usepackage{times}
\usepackage{epsfig}
\usepackage{graphicx}
\usepackage{amsmath}
\usepackage{amssymb}
\usepackage{authblk}
\usepackage{multirow}
\usepackage{bbm}
\usepackage{tablefootnote}
\usepackage{subcaption}
\usepackage{enumitem}
\usepackage{booktabs}

\usepackage[font=small,skip=0pt]{caption}
\usepackage{listings}
\usepackage{xcolor}
\usepackage{fancyhdr}

\graphicspath{{./images/}}

\definecolor{codegreen}{rgb}{0,0.6,0}
\definecolor{codegray}{rgb}{0.5,0.5,0.5}
\definecolor{codepurple}{rgb}{0.58,0,0.82}
\definecolor{backcolour}{rgb}{0.95,0.95,0.92}

\newcommand{\yang}[1]{\textcolor{black}{#1}}

\newcommand{\mathleft}

\lstdefinestyle{mystyle}{
    backgroundcolor=\color{backcolour},   
    commentstyle=\color{codegreen},
    keywordstyle=\color{magenta},
    numberstyle=\tiny\color{codegray},
    stringstyle=\color{codepurple},
    basicstyle=\ttfamily\footnotesize,
    breakatwhitespace=false,         
    breaklines=true,                 
    captionpos=b,                    
    keepspaces=true,                 
    numbers=left,                    
    numbersep=5pt,                  
    showspaces=false,                
    showstringspaces=false,
    showtabs=false,                  
    tabsize=2
}

\cvprfinalcopy % *** Uncomment this line for the final submission

\ifcvprfinal\fi
\begin{document}

\setlength{\abovedisplayskip}{0pt}
\setlength{\belowdisplayskip}{0pt}

%%%%%%%%% TITLE
\title{ReMOTS: Self-Supervised  Refining Multi-Object Tracking and Segmentation\\
($1^{st}$ place solution for MOTSChalelnge 2020 Track 1)}
\author[1,2]{Fan Yang\thanks{Corresponding email: yang.fan.xv6@is.naist.jp}}
\author[1]{Xin Chang}
\author[1]{Chenyu Dang}
\author[3]{Ziqiang Zheng}
\author[1,2]{Sakriani Sakti}
\author[1,2]{Satoshi Nakamura}
\author[4]{Yang Wu}

\affil[1]{Nara Institute of Science and Technology, Japan}
\affil[2]{RIKEN, Center for Advanced Intelligence Project, Japan}
\affil[3]{\yang{UISEE Technology (Beijing) Co. Ltd.}, China}
\affil[4]{\yang{Kyoto University}, Japan}

\maketitle
%\thispagestyle{empty}

%%%%%%%%% ABSTRACT

\begin{abstract}
We aim to improve the performance of Multiple Object Tracking and Segmentation (MOTS) by refinement. However, it remains challenging for refining MOTS results, which could be attributed to that appearance features are not adapted to target videos and it is also difficult to find proper thresholds to discriminate them. To tackle this issue, we propose a self-supervised  refining MOTS (i.e., ReMOTS) framework. ReMOTS mainly takes four steps to refine MOTS results from the data association perspective. (1) Training the appearance encoder using predicted masks. (2) Associating observations across adjacent frames to form short-term tracklets. (3) Training the appearance encoder using short-term tracklets as reliable pseudo labels. (4) Merging 
short-term tracklets to long-term tracklets utilizing adopted appearance features and thresholds that are automatically obtained from statistical information. Using ReMOTS, we reached the $1^{st}$ place on CVPR 2020 MOTS Challenge 1~\cite{Voigtlaender19CVPR_MOTS}, with a sMOTSA score of $69.9$.
\end{abstract}

%%%%%%%%% BODY TEXT
\section{Introduction}
Multiple Object Tracking (MOT)\yang{, which} depends on information from the bounding box\yang{, faces} a great challenge, since different objects may \yang{stay} in the same bounding box and increase the ambiguity to distinguish them. \yang{Recently}, some researchers \yang{in} this filed have moved their eyes to Multiple Object Tracking and Segmentation (MOTS) and hope to take advantage of \yang{object-instance} masks. Under such a background, the first MOTS challenge is organized to explore solutions for MOTS. 
%It is \yang{a great honor} for us to
We participated in this challenge (May-30th-2020) and won the $1^{st}$ place on Challenge 1. In this paper, we represent our solution.

\section{Method Details}

\begin{figure*}[!h]
\captionsetup{font=footnotesize}
\begin{center}
  \includegraphics[width=\linewidth]{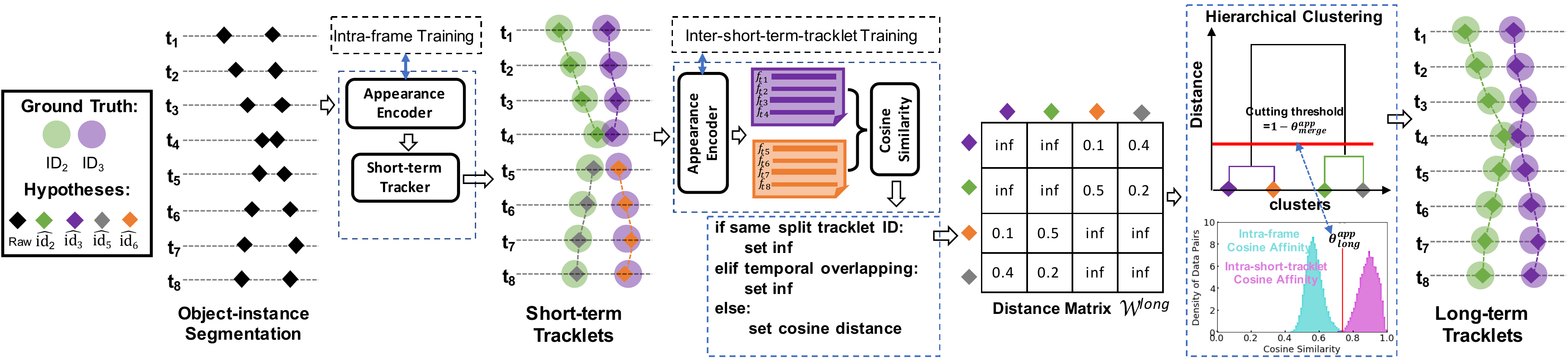}
\end{center}
  \caption{ The illustration of ReMOTS Framework. }
  \label{fig:framework}
\end{figure*}

Overall, we apply \yang{the} tracking-by-detection strategy to generate MOTS results.
Since our ReMOTS is an offline approach, we \yang{refine} the data association by retraining the appearance feature encoder. In each step of ReMOTS, we give a practical guidance to quantitatively select hyperparameters. Our approach is illustrated in \yang{Figure}~\ref{fig:framework}.  After obtaining object-instance masks, we perform: (1) encoder training with intra-frame data, (2) associate masks to short-term tracklets by a short-term tracker, (3) inter-short-tracklet encoder retraining, and (4) merging short-term tracklets.

\begin{figure*}[!h]
\captionsetup{font=footnotesize}
\begin{center}
  \includegraphics[width=0.77\linewidth]{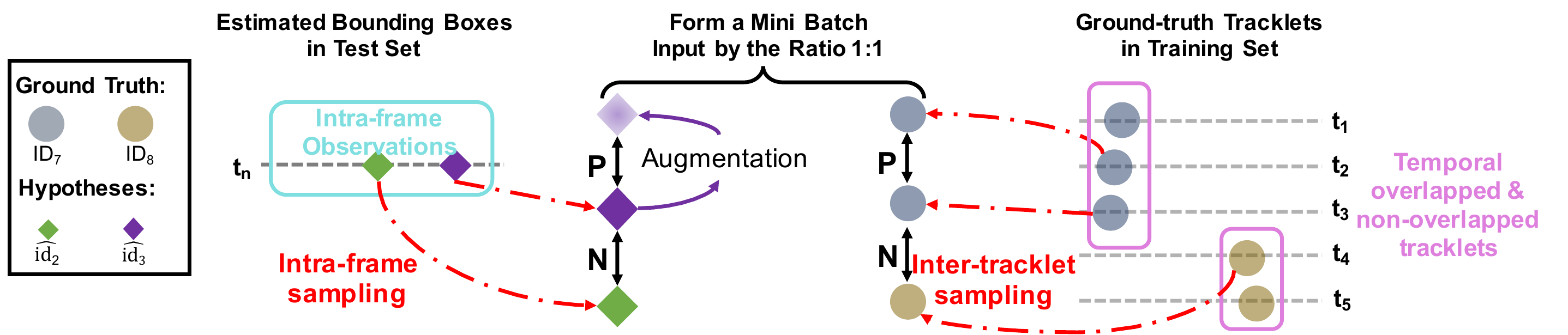}
\end{center}
  \caption{ Constructing training samples for intra-frame training. P and N represent positive and negative samples, respectively. }
  \label{fig:intra-frame}
\end{figure*}

\subsection{Generate \yang{Object-Instance} Masks}
Referring to how the public detection is generated, we obtain \yang{object-instance} masks using the Mask R-CNN X152 of Detectron 2~\cite{wu2019detectron2} and X-101-64x4d-FPN of MMDetection~\cite{mmdetection}. We fuse their segmentation \yang{results} by a modified Non-maximum Suppression (NMS). Unlike the traditional NMS, \yang{where} the IoU (Intersection over Union) is applied, we propose \yang{a new metric named} IoM (Intersection over Minimum) \yang{for it} since heavily overlapped masks may also have low IoU values. The python code of IoM is as follows.
\begin{lstlisting}[language=Python]
def pixel_iom(target,prediction):
    """
    Inputs:
        target: binary mask, array([H,W])
        prediction: binary mask, array([H,W])
    Outputs:
        iom_score: float  
    """
    intersection = np.logical_and(target, prediction)
    min_area = min(np.sum(target),np.sum(prediction))
    iom_score = np.sum(intersection) / min_area
    return iom_score
\end{lstlisting}

After \yang{performing} our modified NMS, the \yang{remaining} masks may still have overlapped areas. Therefore, we only keep the mask with the top confident score \yang{for each overlapping area}.

\begin{figure*}[!h]
\captionsetup{font=footnotesize}
\begin{center}
  \includegraphics[width=0.7\linewidth]{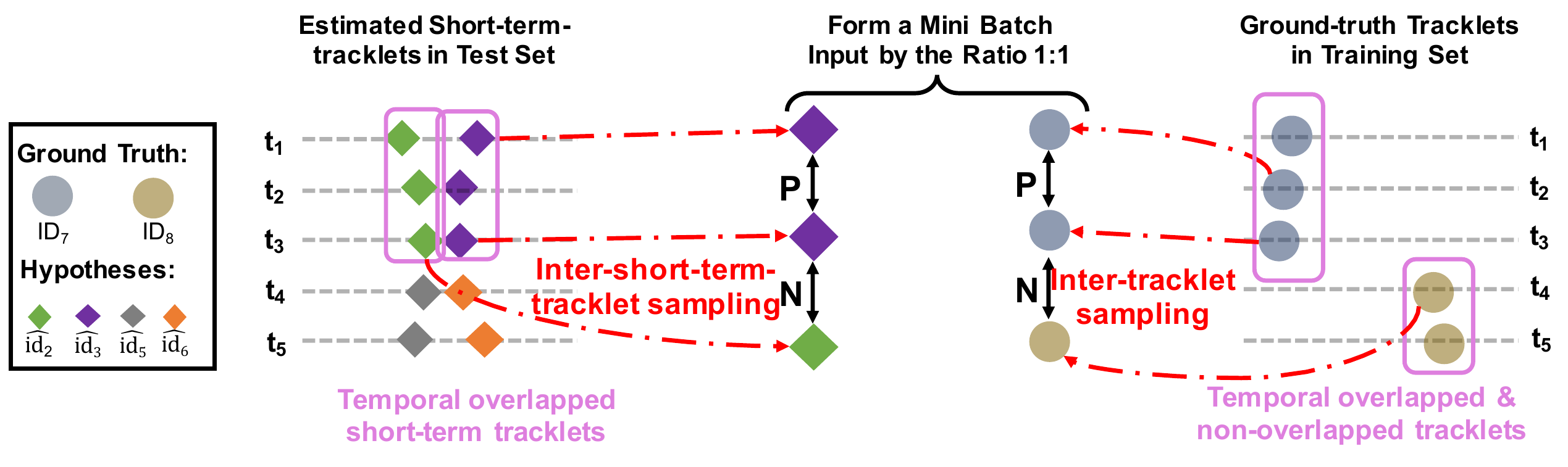}
\end{center}
  \caption{ Constructing training samples for inter-short-tracklet training. P and N represent positive and negative samples, respectively.}
  \label{fig:inter-short-tracklet}
\end{figure*}

\subsection{\yang{Encoder Training with Intra-frame Data}}

We take an off-the-shelf appearance encoder and its training scheme from an object re-identification work~\cite{luo2019bag}. \yang{SeResNeXt50} is used as the backbone and its global-average-pooling output, which is a $2048$-dimension vector, is used as the appearance \yang{features}. The triplet loss~\cite{chechik2010large} is applied to train the appearance encoder. To adapt the appearance feature learning to the target videos, we incorporate intra-frame observations of target videos into a novel offline training process.

As \yang{Figure}~\ref{fig:intra-frame} shows, we can sample triplets from \yang{the training} set only referring to the ground-truth tracklets. In test set, since Non-maximum Suppression (NMS) is performed, we assume that predicted object masks are exclusive within the same frame, and therefore it is easy to form negative pairs \yang{with} intra-frame \yang{observations}. Before tracking, we create a positive sample by augmenting an anchor sample. The augmentation process can dramatically change the pixel content of the anchor sample without altering identity. Finally, we take triplets from the \yang{training} set and target set to form a mini-batch input by the ratio \yang{of} $1:1$. Using such new training samples, we retrain the appearance encoder to obtain more discriminative appearance features.

\subsection{Short-term Tracker}
After intra-frame training, we apply the appearance encoder to generate appearance features for data association. Since the tracker part is not our main focus, we build a simple \yang{tracker} that only associates two-frame observations at once. Using the dense optical flow function of OpenCV, we generate optical flow between two \yang{adjacent} frames, and then warp the mask from previous frame to current frame to calculate IoU of cross-frame masks. The distance matrix \yang{is} formulated as follows:
\begin{equation}
\label{eq:app_1}
%\scalebox{0.9}{
\begin{aligned}
\mathcal{W}^{short}_{prev,curr} &= \\
&\left\{\begin{matrix}
~~~~inf,  ~~~~~if~ IoU(mask_{prev}, mask_{curr})=0 \\ 
1-\frac{f_{prev}f_{curr}}{\left \| f_{prev}\right \| \left \| f_{curr}\right \|},  ~~~~~otherwise
\end{matrix}\right.
\end{aligned}
%}
\end{equation}
where \yang{$mask_{prev}$ and $mask_{curr}$} respectively \yang{denote the mask of the} previous frame \yang{and the mask of the} current frame; $\mathcal{W}^{short}_{prev,curr}$ is their edge weight (i.e., distance); $f_{prev}$ and $f_{curr}$ are their appearance features.

\begin{figure}[!h]
\captionsetup{font=footnotesize}
\begin{center}
  \includegraphics[width=0.5\linewidth]{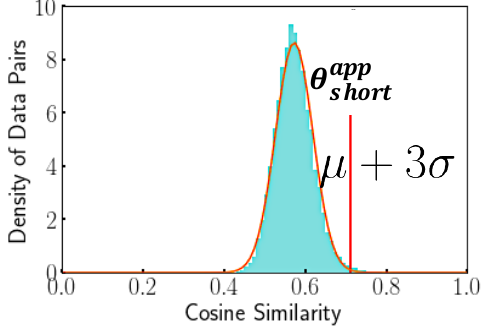}
\end{center}
  \caption{The appearance threshold value for short-term tracking.}
  \label{fig:short_app_thred}
\end{figure}

Besides constraining data association with low IoU values, we also hope to constrain data association with low appearance similarity. However, it is tricky to heuristically determine a threshold \yang{for} constraining. We tackle this issue by analyzing the intra-frame distribution. Specifically, the histogram of appearance cosine similarity between intra-frame masks can be approximated by a normal distribution, and within three standard deviations is $99.7\%$ of the observation pairs (see Figure~\ref{fig:short_app_thred}). We set an appearance affinity threshold at three standard deviations, as value $\theta^{app}_{short}$.

Consequently, using automatically obtained $\theta^{app}_{short}$, we further process $\mathcal{W}^{short}_{prev,curr}$ by
\begin{equation}
\label{eq:app}
\scalebox{0.9}{
\begin{math}
\begin{split}
\mathcal{W}^{short}_{prev,curr} =
\left\{\begin{matrix}
inf,  &if~ \mathcal{W}^{short}_{prev,curr}> 1-\theta^{app}_{short}\\ 
\mathcal{W}^{short}_{prev,curr},  &otherwise
\end{matrix}\right. .
\end{split}
\end{math}}
\end{equation}

We apply linear assignment on \yang{$\mathcal{W}^{short}_{prev,curr}$} to determine the association of masks between \yang{the} previous frame and \yang{the current frame}. Due to \yang{misdetection} and occlusion, such a process can only generate short-term tracklets. However, short-term tracklets reduce the risk of mixing \yang{different} identities, which is an important condition in next-step process.

\subsection{\yang{Inter-short-tracklet Encoder Retraining}}

As we assume each short-term tracklet may only contain a unique identity, they can be used as dependable pseudo labels to train the feature encoder. However, different short-term tracklets, which have no overlap in the temporal domain, may still hold the same \yang{identity}. Therefore, we do inter-short-tracklet retraining under the constraint that sampled short-term tracklets must be temporally overlapping within the same video.

We illustrate the process of training data sampling for inter-short-tracklet training as \yang{Figure}~\ref{fig:inter-short-tracklet} shows. Within a video, we first sample two identities that appear in a \yang{randomly chosen} frame, and then randomly choose another frame for one of the selected identities, thus constructing a triplet. Other settings of inter-short-tracklet training are the same as intra-frame retraining. We update the appearance features after inter-short-tracklet retraining and use them in the next step.

\begin{table*}[h!]
%\captionsetup{font=footnotesize}
\centering
\scriptsize{
\begin{tabular}{c|c|ccccccc}
\toprule
Rank &Method &sMOTSA$\uparrow$  & IDF1$\uparrow$ &MOTSA$\uparrow$  & MOTSP$\uparrow$ & MODSA$\uparrow$	 &MT$\uparrow$ & ML$\downarrow$  \\
\midrule
$1^{st}$ place & ReMOTS (ours) & \textbf{$69.9$} &$75.0$ &\textbf{$83.9$} &$84.0$ &\textbf{$85.1$} &\textbf{$248$} &$12$\\
$2^{nd}$ place & PTPM & $68.8$ &$68.5$ &$82.6$ &$84.1$ &$83.7$ &$244$ &$19$\\
$3^{rd}$ place & PT & $68.4$ &$64.9$ &$82.6$ &$83.9$ &$84.4$ &\textbf{$248$} &\textbf{$10$}\\
\bottomrule
\end{tabular}}
%\vspace{5}
\caption{The performance on CVPR 2020 MOTS Challenge test set (up to submission deadline at May-30th-2020). 
\label{tab:mots20}}
\end{table*}

\begin{table*}[h!]
%\captionsetup{font=footnotesize}
\centering
\scriptsize{
\begin{tabular}{c|c|ccccccc}
\toprule
Sequence &Method &sMOTSA$\uparrow$  & IDF1$\uparrow$ &MOTSA$\uparrow$  & MOTSP$\uparrow$ & MODSA$\uparrow$	 &MT$\uparrow$ & ML$\downarrow$  \\
\midrule
MOTS20-01 & ReMOTS  & $68.5$ &$81.9$ &$83.9$ &$82.5$ &$84.8$ &$8$ &$0$\\
MOTS20-06 & ReMOTS & $74.9$ &$76.8$ &$88.8$ &$85.0$ &$90.6$ &$156$ &$4$\\
MOTS20-07 & ReMOTS & $65.4$ &$68.3$ &$80.6$ &$81.9$ &$81.6$ &$36$ &$4$\\
MOTS20-12 & ReMOTS & $71.8$ &$82.0$ &$82.9$ &$87.2$ &$83.8$ &$48$ & $4$\\
\bottomrule
\end{tabular}}
%\vspace{5}
\caption{The performance of ReMOTS on each sequence of CVPR 2020 MOTS Challenge test set (up to submission deadline at May-30th-2020). 
\label{tab:mots_seq}}
\end{table*}

\subsection{Merging Short-term Tracklets}

With \yang{better} appearance \yang{features} and more robust spatio-temporal information of short-term tracklets,
we are able to merge short-term tracklets into long-term \yang{ones}. The merging process is summarized in \yang{Figure} \ref{fig:framework}. Short-term tracklets association is formulated as a hierarchical clustering problem on a weighted graph, in which each node represents a tracklet and \yang{the graph} edges are represented in a distance matrix $\mathcal{W}^{long}$, defined as
\begin{equation}
\label{eq:app}
\scalebox{0.85}{
\begin{math}
\begin{aligned}
\mathcal{W}^{long}_{k1,k2} = &\\
&\left\{\begin{matrix}
inf,  ~~~~~~if~k1=k2 \\ 
inf,  ~~~~~~if~Distance(\Pi _{k1},\Pi _{k2}) >\theta^{t}\\ 
inf,  ~~~~~~if~ \Pi _{k1} \cap \Pi _{k2}\neq \varnothing \\ 
\frac{1}{N_{k1}N_{k2}}
\sum_{i \in \Pi _{k1}}
\sum_{j \in \Pi_{k2}}
\big(1-\frac{f^{k1}_{i}f^{k2}_{j}}{\left \| f^{k1}_{i}\right \| \left \| f^{k2}_{j}\right \|} \big),   otherwise
\end{matrix}\right. ,
\end{aligned}
\end{math}}
\end{equation}
where for tracklets $T_{k1}$ and $T_{k2}$, $\mathcal{W}^{long}_{k1,k2}$ is their edge weight (i.e., distance); $\Pi _{k1}$ and $\Pi _{k2}$ are their temporal ranges; $f^{k1}_{i}$ and $f^{k2}_{j}$ are their appearance features at frame $i$ and $j$, and $N_{k1}$ and $N_{k2}$ are the number of observations within the tracklets, respectively.

Whenever the matching condition between two short-term tracklets violates any of the following three principles: (1) different short-term track ID, (2) the temporal gap between two short-term tracklets are within $\theta^{t}$ frames (we use $\theta^{t}=15$ ), and (3) no temporal overlap between two short-term tracklets, we set their distance value to be infinite. To hold these constraints in the whole process of hierarchical clustering, we apply the centroid \yang{linkage} criteria to determine the distance between clusters. 

\begin{figure}[!h]
\captionsetup{font=footnotesize}
\begin{center}
  \includegraphics[width=0.6\linewidth]{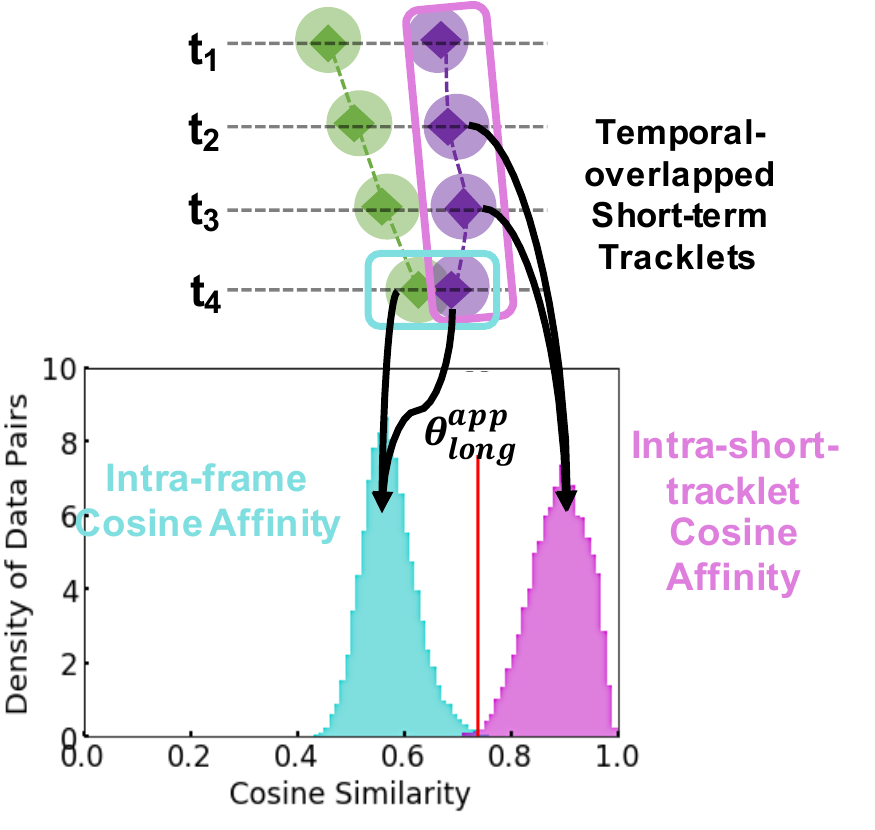}
\end{center}
  \caption{The appearance threshold value for merging short-term tracklets.}
  \label{fig:long_app_thred}
\end{figure}

The main challenge of applying hierarchical clustering is \yang{on} how to set \yang{a} proper cutting threshold. We do not give a heuristic value, \yang{and we} let the data speak for themselves \yang{instead}. We suppose that intra-frame and inter-short-tracklet cosine similarity histograms can be separated at $\theta^{app}_{long}$ (see Figure~\ref{fig:long_app_thred}) after inter-short-tracket retraining, though small overlapping might \yang{exist}. Without accessing to the ground-truth, this could be a reasonable boundary to distinguish objects based on appearance features. Therefore, we set $1-\theta^{app}_{long}$ as the cutting threshold in hierarchical clustering.

\subsection{Experimental Setup}
The only neural network model - appearance encoder~\cite{luo2019bag}, used in this work, is not our contribution and our ReMOTS can do the same \yang{refining when other appearance models are used}. Therefore, we do little change \yang{to} the default setting of \cite{luo2019bag}, except for forming novel training samples in our intra-frame training and inter-short-tracklet training. Here, we omit \yang{the} other details described in \cite{luo2019bag}.

\section{Results}
We report the performance of our ReMOTS on the MOTChallenge evaluation system, with metrics introduced \yang{in} \cite{Voigtlaender19CVPR_MOTS}. In Table~\ref{tab:mots20}, we list the performance of top-3 methods up to the submission deadline. Our method mainly outperform \yang{the other} two methods in terms of IDF1 score, and therefore leads to state-of-the-art performance in this challenge. The detailed performance \yang{on} each test sequence \yang{is} listed in Table~\ref{tab:mots_seq}. Though the same method is applied, it can be observed that \yang{the performance of each sequence varies a lot}. This may be attributed to the diversity between videos, which calls for more exploration in automatically adapting MOTS models to target videos. Our ReMOTS analyzes the statistical information at the entire video level, but the temporal local statistical information, which might be \yang{useful} for \yang{fine-grained} adaption, has not been considered yet.

\section{Conclusion}
We present our \yang{solution which wins} the CVPR 2020 MOTS Challenge 1. In our proposed ReMOTS framework, intra-frame training and inter-short-tracklet training are introduced for learning better appearance features for \yang{more effective data association}, which are our main contributions. Besides, we quantitatively demonstrate how to select proper thresholds by analyzing the statistical information of tracklets, which could be useful for other multiple object tracking works. The main limitation of ReMOTS is that it cannot be used in real-time scenarios, but it may bring insights to design better online MOTS method with \yang{feature adaptation}.

\section*{ACKNOWLEDGEMENTS}
This work was supported by JSPS KAKENHI Grant Numbers JP17H06101.

{\small
\bibliographystyle{ieee_fullname}
\bibliography{egbib}
}

\end{document}